\documentclass[sigconf]{acmart}
\usepackage{booktabs} 
\usepackage{multirow}

\setcopyright{rightsretained}

\acmConference[ACM]{ACM Multimedia conference}
\acmYear{2018}
\copyrightyear{2018}

\begin{document}
\title{X-GANs: Image Reconstruction Made Easy for Extreme Cases}

\author{Longfei Liu}

\affiliation{%
  \institution{Peking University \& KingSoft}
}

\author{Sheng Li}
\authornote{corresponding author: lisheng@pku.edu.cn}
\affiliation{%
  \institution{Peking University}
}

\author{Yisong Chen}

\affiliation{%
  \institution{Peking University}
}

\author{Guoping Wang}

\affiliation{%
  \institution{Peking University}
}


\begin{abstract}
Image reconstruction including image restoration and denoising is a challenging problem in the field of image computing. We present a new method, called X-GANs, for reconstruction of arbitrary corrupted resource based on a variant of conditional generative adversarial networks (conditional GANs). In our method, a novel generator and multi-scale discriminators are proposed, as well as the combined adversarial losses, which integrate a VGG perceptual loss, an adversarial perceptual loss, and an elaborate corresponding point loss together based on the analysis of image feature. Our conditional GANs have enabled a variety of applications in image reconstruction, including image denoising, image restoration from quite a sparse sampling, image inpainting, image recovery from severely polluted block or even color-noise dominated images, which are extreme cases and haven't been addressed in the status quo. We have significantly improved the accuracy and quality of image reconstruction.  Extensive perceptual experiments on datasets ranging from human faces to natural scenes demonstrate that images reconstructed by the presented approach are considerably more realistic than alternative work. Our method can also be extended to handle high-ratio image compression. 
\end{abstract}

%
%


\maketitle

\section{Introduction}

Image reconstruction is an important issue in the field of image computing, which covers a broad scope including image restoration, image inpaiting, image denoising, image super-resolution, etc. It has been widely investigated for a long period. Deep convolutional networks \cite{he2016deep,krizhevsky2012imagenet,simonyan2014very} made great progress in recent years in the field of computer vision. Nowadays, as deep learning has been widely used, a general idea may arise to let the machine learn how to reconstruct images with deep learning. The advantage is that the image can be reconstructed through learning from a large amount of samples without specific analysis for different types of image recovery respectively. Many good results have been achieved using this strategy
\cite{dong2016image,levin2004colorization,larsson2016learning,cho2016natural,yeh2017semantic,levin2008closed,burger2012image}.

\begin{figure}[]
\includegraphics[width=1.65in]{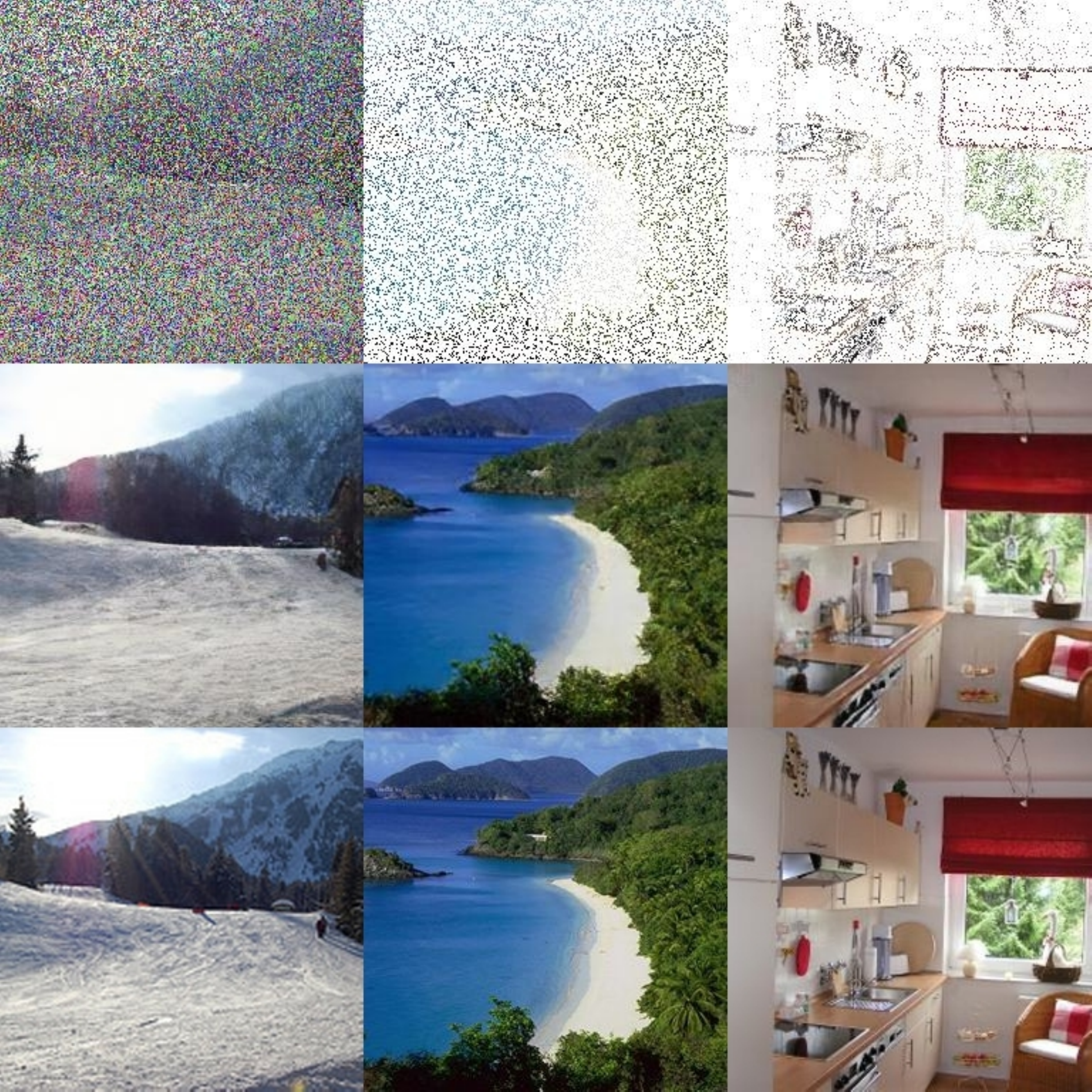}
\includegraphics[width=1.65in]{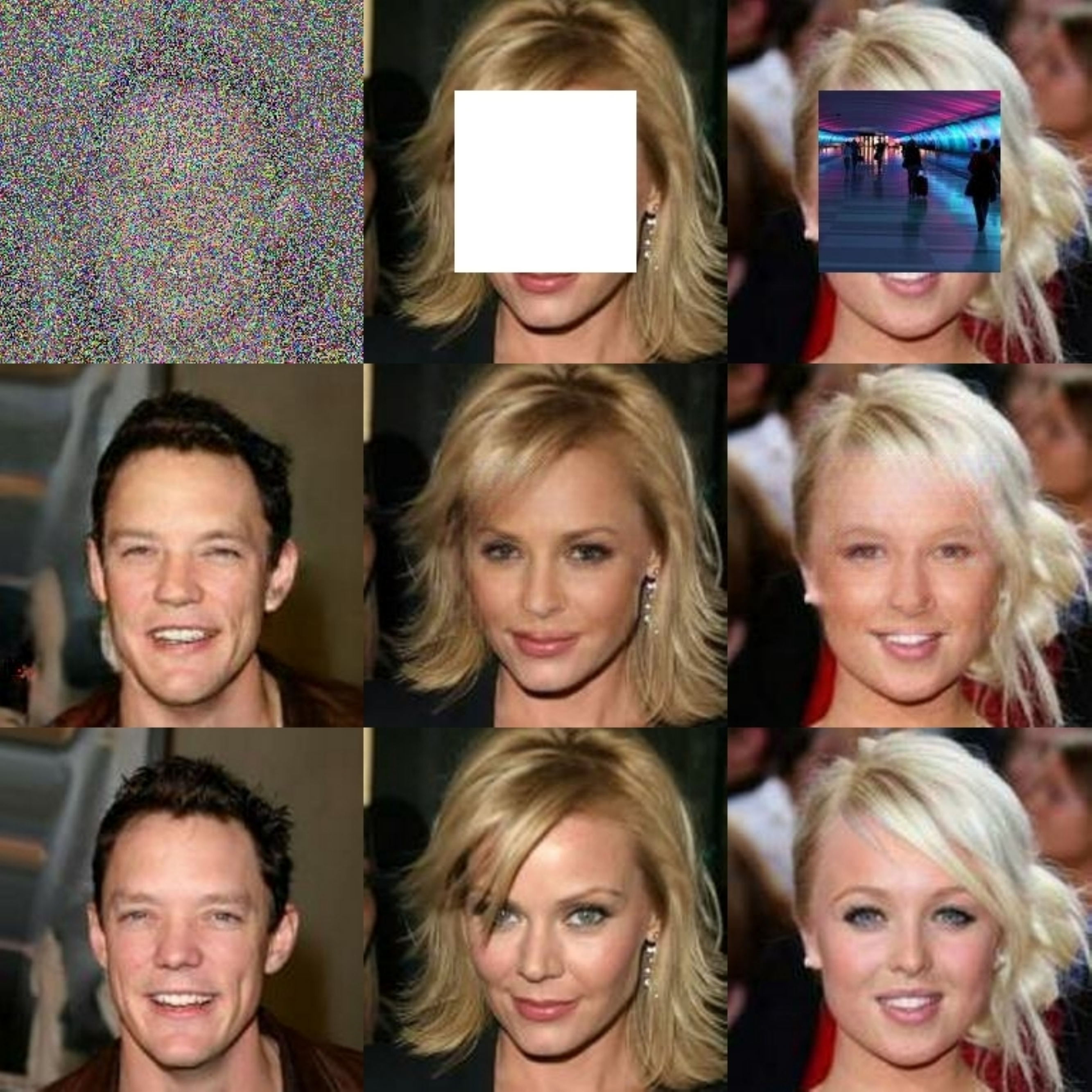}
\caption{Various benchmarks for image reconstruction. The first row shows the corrupted images, the second row shows the reconstructions by our method, and the third row shows the real images accordingly. }
\end{figure}

Generative adversarial networks (GANs) \cite{goodfellow2014generative} have brought new ideas for deep learning since 2014. GANs are a framework for producing a generative model by way of a two-player minimax game. Although it has succeeded in a lot of practical applications, such as synthesizing interior decorations, human faces, and street scenes, it is not an easy task for generative adversarial networks to obtain high-quality training samples because a lot of work has to be spent on time-consuming manual annotation, which is a common drawback of modern deep convolutional neural networks.

There are many developments in image reconstruction using generative adversarial networks, such as super-resolution \cite{dong2016image}, matting\cite{cho2016natural,levin2008closed}, inpainting \cite{pathak2016context,yeh2017semantic}, colorization \cite{levin2004colorization,xu2013sparse,larsson2016learning}, denoising \cite{burger2012image,liu2016learning}. Because a large number of high-quality images can be obtained as training samples, we can easily train a high-quality network model.

To recover image from as few pixels as possible in the presence of moderate noise is the goal of our work. We aim to use the distribution of least pixels and the 'imagination' with the GANs to reconstruct the image to the maximal extent. This can be a challenge of image reconstruction, and definitely is of great significance in CT reconstruction, image compression and transmission, and so on.

A common problem of GANs network is that it can only handle relatively small images due to network capacity. Therefore, a more stable and faster training method is required for larger images, because the network training may become more difficult to converge when the size of image increases. Furthermore, unfavorable results may occur when dealing with more complex conditions, such as too much information missing in the source images. It seems impossible to recover the image in the severe absence of original information. Under such condition, we still want to obtain a stable result that is consistent with people's perception, both at the coarse level of overview and the fine level with local detail. This puts high demands on our network structure and training methods.

The image reconstruction framework addressed in this paper mainly includes the following aspects:

Firstly, only a small number of discrete samples are used to reconstruct the original image, also called image interpolation. Unlike Ruohan Gao et al.'s goal to reconstruct 64x64 or 128x128 images with more than 25\% discrete points \cite{gao2017demand}, we try to challenge even harder problem where more than 80\% of a 256x256 image are corrupted. Faced with the problem of restoring a small amount of random discrete points, the predecessors have done many useful explorations, such as the use of nonparametric probabilities, density function estimation methods, and other attempts. Previous papers \cite{gao2017demand,yeh2017semantic} solved the problem to some extent. In order to pursue the use of fewer known discrete points and achieve better and more stable results, we consider using an improved counter-neural network to solve this problem. We re-assembled the network model, combining some of the previous experience. We have chosen some more appropriate network frameworks and modified the network so that our network can restore the original image better. Experimental results show that that neural networks have their own unique advantages in solving such problems. After learning a large number of data samples, we are able to associate the mapping relationship between discrete points and corresponding images. This form of GANs network has also achieved good results on recovering problems using very few discrete points (discrete points less than 20\%).

Secondly, in order to further compress the number of known discrete points, we consider that scattering discrete points in a specific area can benefit image restoration. So we use the sobel \cite{sobel2014history} or canndy operator  to extract the edge of the image, in order to distribute the discrete points in positions around high gradients. Our experiments discovered that using the Sobel operator to specify images with discrete points can better restore the image at the same percentage of sample points, and restore the details and edges of the image more clearly, which accords with the human's visual system.

Finally, we design more complex scenarios to challenge the ability of network training. We try to recover from images with random colored noise (no longer a white or gray noise background) or from images with cluttered blocks in the original image. These extremely complex situations further increase the difficulty of network design and training. Because the network needs to peel off interference information in a large number of color points to find useful point information that constitutes the original image and the image can be restored from it. We can even handle such complex situation that the correct information from the source is less tan 20\%.

Our main contributions are as the follows:
(1) We propose X-GANs network, which using multi-dimensional loss functions architecture to counteract the problem of missing or obfuscated images and blocks of random discrete points in the network, and achieve better result than before;
(2) The loss function is carefully designed for the problem of complex discrete point picture restoration with color noise interference, which improves the network's ability to recover discrete points, even with very little original image information. The image can be reconstructed with high quality;
(3) By specifying the distribution of discrete points, the bound of network resilience can be explored. This allows the network to recover images with fewer discrete samples. This method also helps in high-ratio compression and transmission of images or videos.

\section{Related Work}
Reconstructing missing image parts \cite{pathak2016context, yeh2017semantic, gao2017demand} has long been a focus in the field of image computing. Early approaches mainly use traditional image processing techniques to restore images from different level of degradation. In recent years. as deep neural networks become widely studied, many attempts based on deep neural networks are made to do image restoration. Among these approaches, the performance of generative adversarial network image restoration is particularly impressive. Therefore, GANs have become an attractive solution to solve related problems.

\subsection{Image Reconstruction}

Image restoration from discrete sampling points is an important reconstruction problem. The noise to be removed is in the form of the same color, or with random arbitrary colors. Both cases can be understood as removing noise of a corrupted image to restore the corresponding real image. We corrupt real images by adding noise drawn from a zero-mean normal distribution with variance s (the noise level). Many related work have been done, such as \cite{gao2017demand, pathak2016context,yeh2017semantic,burger2012image,liu2016learning}, etc. The main difficulty in recovering from discrete point images is when the size of the restored image is too large, or when the number of discrete points is too small, the noise image is difficult to restore back to the ideal effect. Generally, when the image size is larger than 128, the existing network tends to cause problems such as unstable image generation and unclear image detail generation. When the discrete points are less than 25\%, there are often too few discrete points in the image, resulting in the inability to find relationships between them during the training process. It is impossible to extrapolate the relationship between the discrete points in the global perspective. In the case where the discrete points can no longer generate a unique definite image, and when the noise is of random color, the difficulty of using the network to recover the image becomes greater, and the discrete points of the original image will be dispersed in the color noise. Therefore, the network needs a larger receptive field to find useful discrete points and their associated information in the noise, and to assemble it back to the original image.

Since the conditional GANs was proposed \cite{mirza2014conditional}, many improved conditional GANs have attempted to solve such problems as image inpainting. However, the training of the network for such problems is generally not easy, because the size and position of the mask patches are usually not fixed. Perhaps it is easier for the network to find the areas to be reconstructed (usually the isolated areas with a pure color), but it is not easy to recover the missing areas in combination with the existing information. Many related works have been done  \cite{gao2017demand,pathak2016context,yeh2017semantic}, and left some prevalent problems. On one hand, it is difficult to train ideally for images bigger than 128. On the other hand, the occluded parts of the input image are not easily restored with high quality. An ideal network should not only be able to restore larger-size images, but also be able to retain as much of the occluded parts of the input image as possible. In addition, to the best of our knowledge, the problem of image inpainting is generally to restore the image of a blocked patch of a certain color. We further improved the difficulty of the problem by replacing the solid color patches with random images and then to perform image restoration.

\subsection{GANs}
GANs is a framework for training generative parametric models \cite{goodfellow2014generative}, and have been shown to produce high quality images \cite{iizuka2017globally, wang2017high, levin2004colorization}. This framework trains two networks with a generator $G$ and a discriminator $D$. $G$ maps a random vector z, sampled from a prior distribution $p_{\rm{z}}$, to the image space while $D$ maps an input image to a likelihood. The purpose of $G$ is to generate realistic images, while $D$ plays an adversarial role, discriminating between the image generated from $G$, and the real image sampled from the data distribution $p$ data .
The $G$ and $D$ networks are trained by optimizing the following loss function $V\left ( D,G \right )$:
\begin{align}
\underset{G}{min}\, \underset{D}{max}\, V\left ( D,G \right )=\mathbb{E}_{{\rm{x}} \sim p_{data} \left ( \rm{x} \right )}\left [ \log D\left ( \rm{x} \right ) \right ]+ \nonumber\\
\mathbb{E}_{{\rm{z}} \sim p_{\rm{z}}\left ( \rm{z} \right )}\left [ \log \left ( 1-D\left ( G\left (\rm{z} \right ) \right ) \right ) \right ],
\end{align}
where x is the sample from the $p_{data}$ distribution; z is a random encoding on the latent space. With some user interaction, GANs have been applied in interactive image editing \cite{zhu2016generative}. However, GANs can not be applied to the the problem of inpainting directly, because they produce an entirely unrelated image with high probability, unless constrained by previously given corrupted image.

\section{X-GANs network}

\subsection{Pix2pix Baseline}
The pix2pix method \cite{isola2017image} is a conditional GANs framework for image-to-image translation. It consists of a generator and a discriminator. Unlike the normal GANs, the conditional GANs associate the input image with the generated image and input the discriminator together. It allows the network to learn the correlation between the input image and the output image. For our task, the objective of the generator is to translate the discrete point map to realistic-looking images, while the discriminator aims to distinguish real images from the translated ones. The framework operates in a supervised setting. In other words, the training dataset is given as a set of image pairs. For each pair, one image is a discrete point map and the other is the corresponding real photo. Conditional GANs aim to model the conditional distribution of real images given the input discrete point map via a minimax game.

\subsection{Architecture}
Our network consists of a generator and a multi-scale discriminator with multi-dimensional loss functions. These loss functions include an adversarial loss, a VGG perceptual loss, an adversarial perceptual loss, and a customized corresponding point loss. The target image is obtained by the generator, the input image is then concatenated with the target image and the real image respectively, and the discriminator network is used to calculate adversarial perceptual loss and adversarial loss. After that, the target image and the real image are calculated by the perceptual loss through the VGG network. Then the input image is used to mask the target image, and the result is compared with the input image to calculate the L2 loss value. Finally, several losses are weighted and are back propagated to the generator.

\begin{figure}
\includegraphics[width=3.2in]{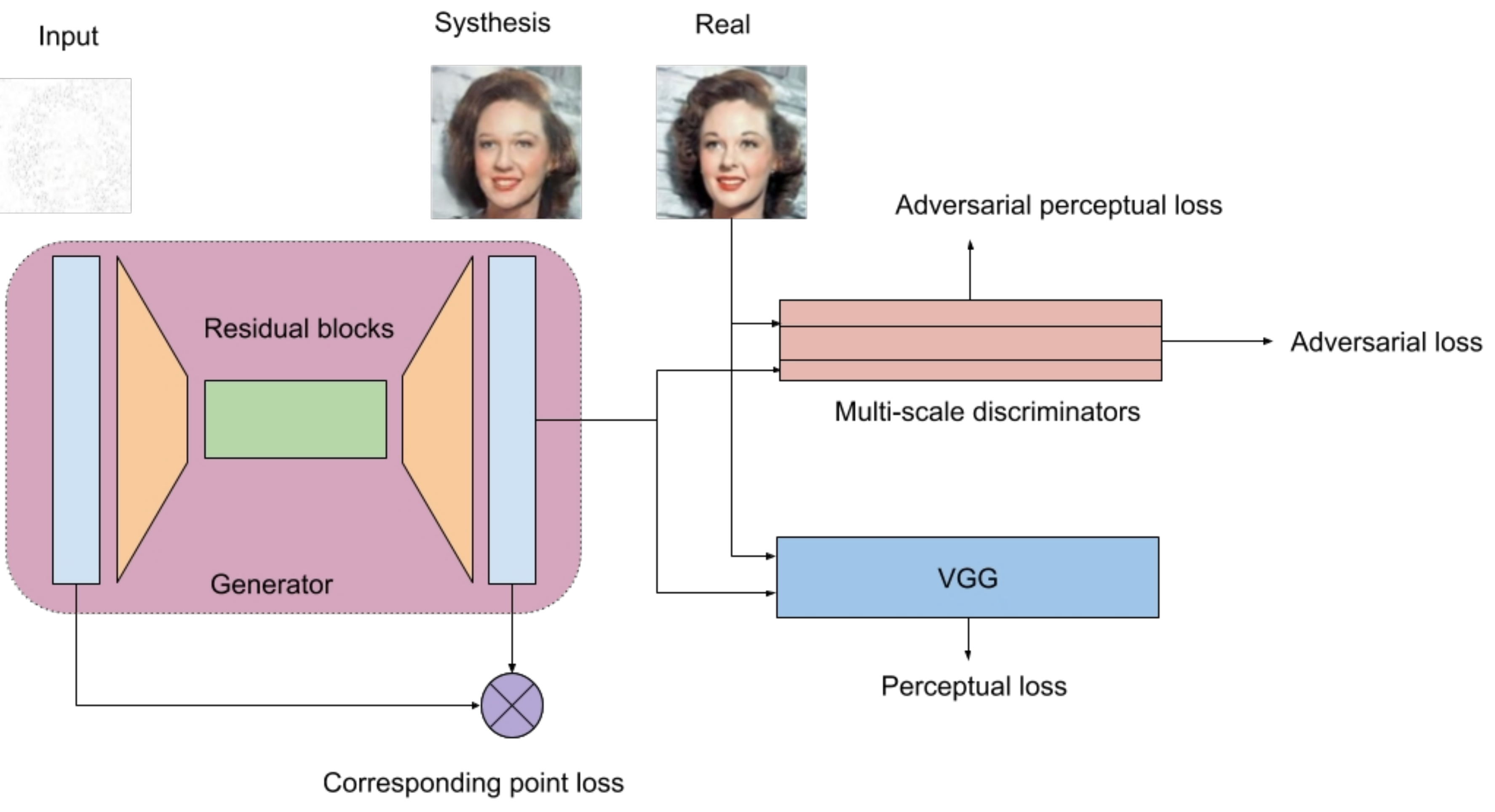}
\caption{Framework of the our network using GANs. We use a multi-dimensional loss functions architecture to help improve the quality of image reconstruction.}
\end{figure}

\subsection{Generator}
Generator is an important part of the network. Because the image to be reconstructed is generally obtained by the generator, it directly determines the final quality of image reconstruction. It is the main challenge to recover high-quality image with fast training speed and preserve the details of the images. In the selection of generators, we refer to the work by Johnson et. al \cite{johnson2016perceptual}. Because this network has been proven to be effective in the generation of large-scale images. This network is composed of downsampling \cite{long2015fully,noh2015learning}, residual blocks \cite{he2016deep}, and upsampling \cite{long2015fully,noh2015learning}. The use of residual blocks, which refers to the residual learning and batch normalization \cite{ioffe2015batch}, can greatly benefit the CNN learning as it can not only speed up the training but also boost the denoising performance \cite{zhang2017beyond}. So we design the structure of the multiple blocks to link the downsampling modules and the upsampling modules as in \cite{johnson2016perceptual}, and both the downsampling and the upsampling modules use a deconvolution structure without the unpooling module. It has such benefit as the relative low computational cost and the effective receptive field sizes \cite{johnson2016perceptual}, each of which contains two 3x3 convolutional layers. We use three convolution -Batchnorm-ReLu to get smaller size feature map, then use nine residual blocks to capture feature representations. We take the Batchnorm layers and Leaky ReLU after both the convolution operation and the deconvolution layers. The final outputs have been normalized to the interval [-1,1] using the Tanh activation to generate a 256x256 output image.


\subsection{Multi-scale Discriminators}
GANs are reputably difficult to train. Especially, for the reconstruction of large-scale image with few known discrete points, the stability of GANs should be guaranteed. In order to solve this problem, we refer to the structure in \cite{durugkar2016generative}. We use 3-discriminator extension to the GANs framework that have an identical network structure to process the image. The generator trains using feedback aggregated over multiple discriminators. If $F := max$ , $G$ trains against the best discriminator.High-resolution image synthesis poses a great challenge to the GANs discriminator design \cite{wang2017high}. For random discrete point images (especially when the number of discrete points is small) the network requires a large receptive field. In this way, the network can find the relationship between discrete points that are far away from each other. In order to increase the receptive field of the network, deeper networks or larger convolution kernels are usually needed, but doing so often increases the instability of network training. Based on the above reasons, we use multi-scale discriminators. Multi-scale discriminators have been proved to be able to deal well with large-size image problems in the pix2pixHD network. The discriminators $D1$, $D2$ and $D3$ are then trained to differentiate real and synthesizing images at 3 different scales, respectively. We treat multiple $D$s for images of different sizes so that the network does not need to add more layers, and in the case of a wider convolutional core, the goal of increasing the network receptive field can be achieved.
\begin{equation}
\underset{G}{min}\underset{D1,D2,D3}{max}\sum_{k=1,2,3}\mathcal{L}_{\rm{GAN}}\left ( G,D_{k} \right )
\end{equation}.

\subsection{Loss Functions}

\textbf{Adversarial matching loss:}
To allow for generators to produce natural statistics at multiple scales discriminator, we add the feature matching loss to the overall loss function \cite{wang2017high}. The feature matching loss is processed similarly to the perceptual loss \cite{johnson2016perceptual,dosovitskiy2016generating,gatys2016image}, which has been shown to be useful for image super-resolution \cite{ledig2016photo} and style transfer \cite{johnson2016perceptual}. Specifically, feature matching loss extracts features from multiple layers of the discriminator, and learns to match these intermediate representations between the real image and the synthesized image.
\begin{equation}
\mathcal{L}_{\rm{FM}}\left ( G,D_k \right )=\mathbb{E}_{\left ( \rm{s},\rm{x} \right )} \sum_{i=1}^{T} \frac{1}{N_i} \left [ \left \| D_{k}^{\left ( i \right )}\left ( \rm{s},\rm{x} \right )-D_{k}^{\left ( i \right )}\left ( \rm{s},G\left ( \rm{s} \right ) \right ) \right \|_{1} \right ]
\end{equation}
It denotes the feature value of the i-th layer in the k-th $D$ of the real image, and then subtracts this value from the generated image, and calculates the L1 value.

\begin{equation}
\underset{G}{min}\left ( \left ( \underset{D1,D2,D3}{max}\sum_{k=1,2,3}\mathcal{L} _{\rm{GAN}}\left (G,D_{k}  \right ) \right )+\lambda \sum_{k=1,2,3}\mathcal{L}_{\rm{FM}}\left ( G,D_{k} \right )  \right )
\end{equation}
Get the largest $D$ among multiple $D$s and sum them up with feature matching loss, and then figure out the value $G$ when the smallest value is achieved. In other words, let the $G$ have the smallest value.

\textbf{VGG perceptual loss:}
When comparing GANs generated images with real images, despite their perceptual similarity they are actually very different as measured by per-pixel losses. \cite{johnson2016perceptual} shows that training with a perceptual loss allows the model to reconstruct fine details and edges. So, we use VGG19 as a kind of perceptual loss. The specific approach is to train the generated image and the real image through the VGG network, and take out the results of each dimension to calculate the loss difference \cite{johnson2016perceptual,dosovitskiy2016generating}. The VGG network is a widely used network for region detection. The loss in each layer of the VGG19 represents the details of different properties of the image such as shape, color, texture, etc. The network can learn the image by using various perceptual loss of the VGG network. Multiple dimensions of information make the overall effect of reconstruction more realistic.
We define
\begin{equation}
 \lambda \sum_{i=1}^{N}\frac{1}{M_{i}}\left [ \left \| F^{\left ( i \right )}\left ( \rm{x} \right )-F^{\left ( i \right )}\left ( G\left ( \rm{s} \right ) \right ) \right \|_1 \right ]
\end{equation}
as our objective, where  $\lambda$ represents a scale factor, $F^{\left ( i \right )}$ denotes the $i$-th layer with $M_i$ elements of the VGG network, x represents the real image, s represents the input of the generator, and $G\left ( \rm{s} \right )$ denotes the image generated from the generator.

\begin{figure}
\includegraphics[width=2.9in]{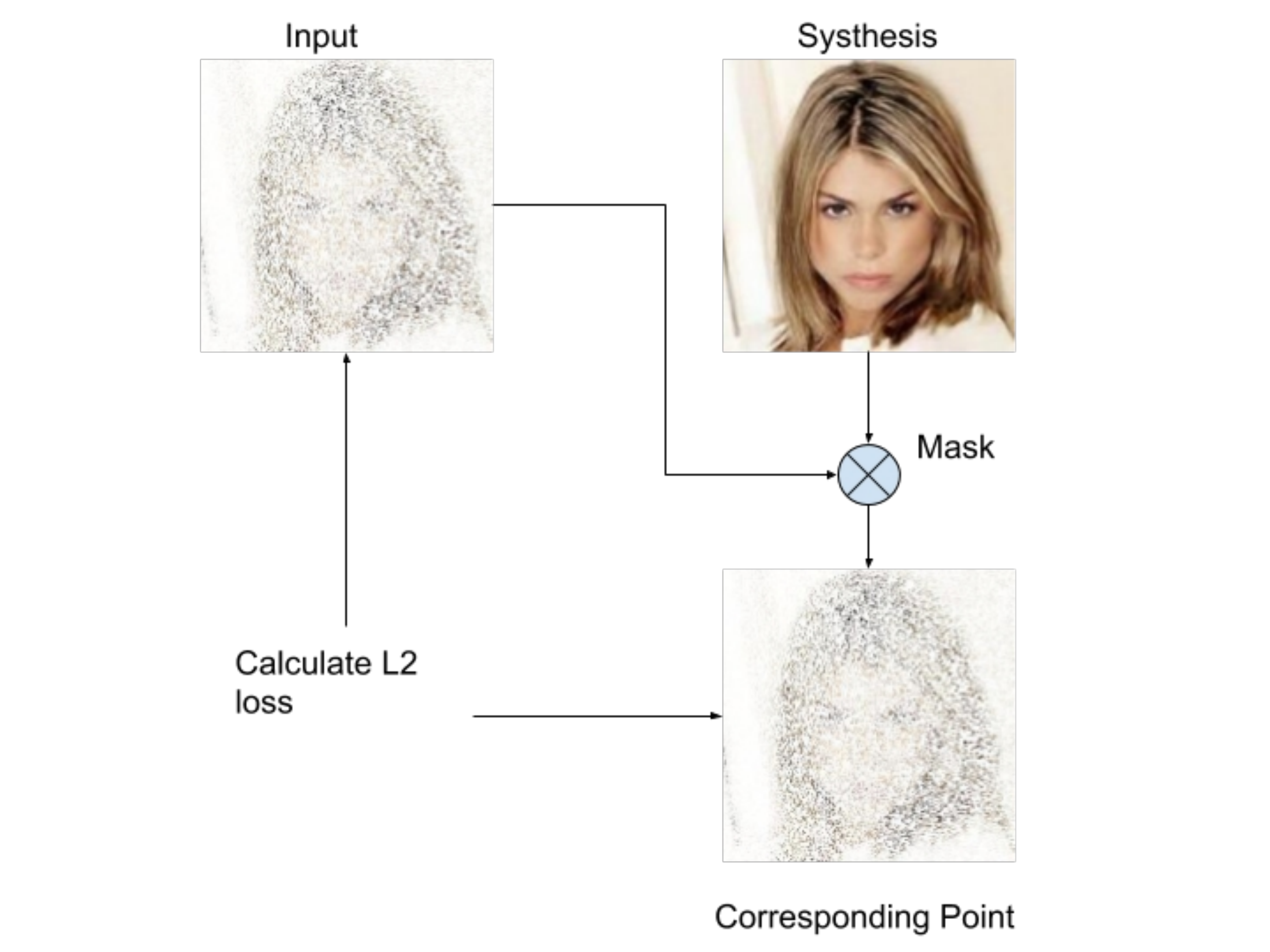}
\caption{Illustration of point loss. It is used in the reconstruction of image inpainting or interpolation in a pixel-to-pixel way.}
\label{fig:pointloss}
\end{figure}

\textbf{Corresponding point loss:}
Different numbers of discrete points often lead to significantly different qualities of restoration. Fewer discrete points generally make image reconstruction more difficult. On one hand, the fewer the discrete points, the smaller the loss value of L2 becomes. In other words, unsuitable L2 loss may lead to excessive punishment for the images with fewer discrete points; on the other hand, the discrete points from the source image should appear in the corresponding position of the reconstructed image. We hope that the neural network can learn the mapping relationship between the input image and the generated image \cite{isola2017image}. In view of the above reasons, we changed the L2 loss function of GANs from calculating the loss value of the generated image with the target image to computing the L2 loss value between the discrete point x and the generated image with mask $G\left ( \rm{s} \right )$ as follows:

\begin{equation}
\left \| \mathrm{s}-Mask\left ( G\left ( \rm{s} \right ),\rm{s} \right ) \right \|_{2}^{2}.
\end{equation}

The workflow of the L2 loss for reconstruction can be seen in Fig. \ref{fig:pointloss}. This can solve the problem of unbalanced punishment, and also solve the problem that the target image does not contain the discrete points of the source image. Through experiments, we find that the use of the new L2 loss function on the basis of the original network is better than without this loss function, and the improvements on PSNR can be observed at each level. Adaptive L2 loss has obvious advantages on image generation where there is a certain correlation between the source image and the target image.

\section{Experiment result}

In our experiment, two data sets are used to train the network, CelebFaces Attributes(CelebA) \cite{liu2015deep} and SUN397 Scenes \cite{xiao2016sun}. All images are resized to 256x256 resolution. CelebFaces Attributes is a face data set. It tests the network's ability to reconstruct a single type of scene. CelebA contains 10,177 identities, and a total of 202,599 pictures. We train 100,000 pictures and use another 500 for verification. SUN397 Scenes is a complex data set of various scenes. It contains 108,754 images in 397 categories. Similarly, the training network is conducted using 100,000 images and 500 proofs were used.

We explore the network's ability to restore images of different scenes through this dataset.
When performing random discrete point reconstruction, we create a 256x256 integer matrix with a discrete uniform distribution in the range [0,99], then set the assigned percentage of entries bigger than a threshold to be white. The generation of discrete points in terms of \cite{sobel2014history} is to obtain an edge detection map by employing a Sobel operator. We calculate the probability of occurrence of each point based on this edge detection map, generate a probability map, and finally select the specified point in the original image based on this probability map and the number of discrete points required. The method for generating the color noise is similar to the generation of random discrete point or the salt-pepper noise, except that the found positions are set to random RGB colors. The white block is used to cut a 128x128 square block in the middle region of the CelebA image. The color blocks are generated by randomly tailing 16x16 to 128x128 color blocks from SUN397 and fill in the data generated by the training and validation set of CelebA. We enable corresponding point loss only for reconstruction of discrete point and white block task, while disable it for color discrete points and color block task. We train our networks with a batch size 1 for 100,000 iterations, giving 50 epochs over the training data. We use Adam\cite{kingma2014adam} with a learning rate of 2x$10^{-4}$. We use dropout to train out generator with 0.5 probability of an element to be zeroed. Training takes roughly 6 days on a single Tesla P40 GPU.
More experimental results can be found in the supplemental material.

\subsection{Recovery from discrete points}
\begin{figure}
\includegraphics[width=3.4in]{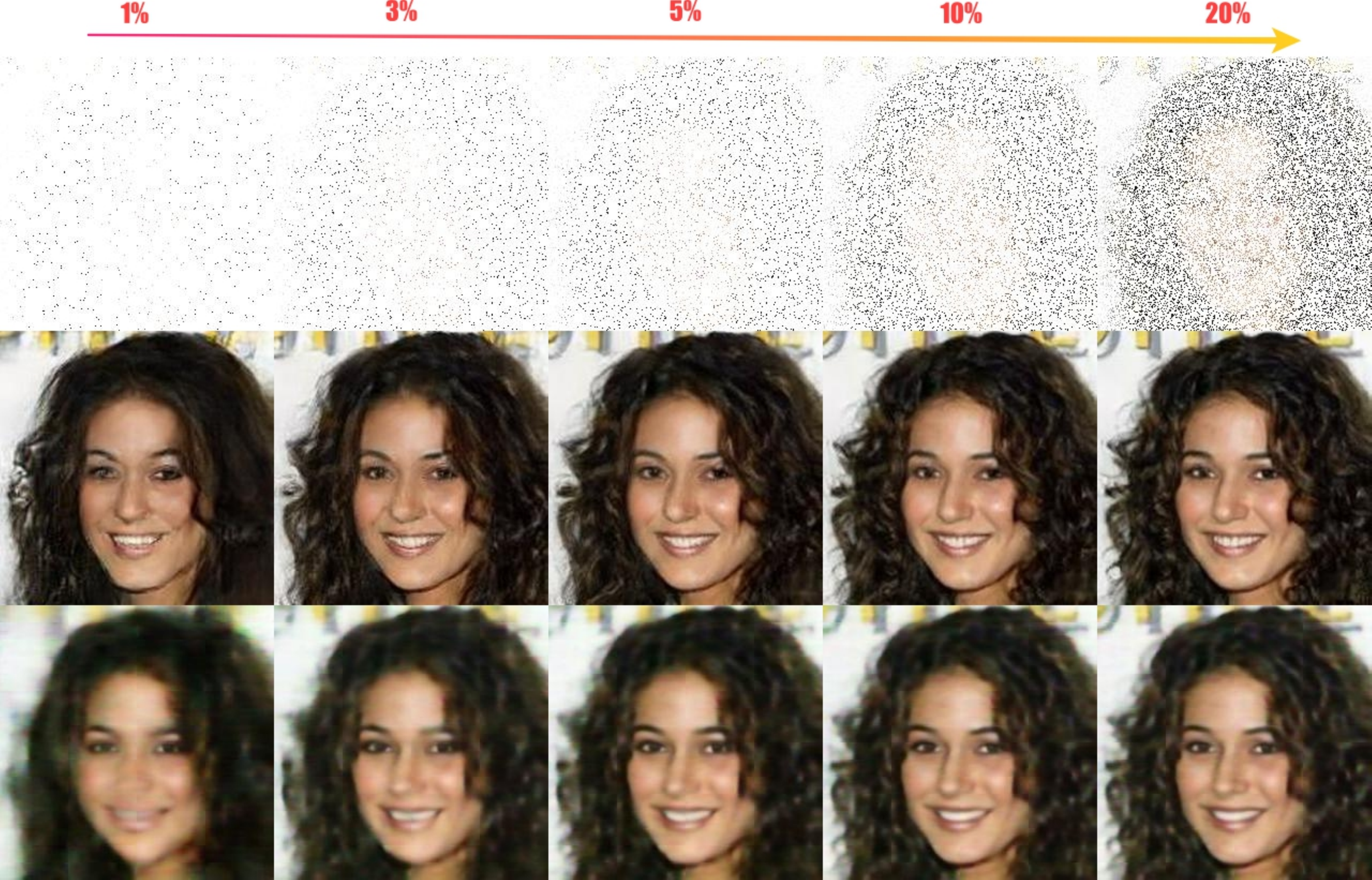}
\caption{Image reconstruction from varying ratio of discrete samples from 1\% $-$ 20\%. The first row is the discrete samples as the source image, the second row is generated by our method, and the third row is generated by \cite{gao2017demand}. Our method can synthesize better results.}
\label{fig:whiteratio}
\end{figure}

For the image restoration of random discrete points, our experiment mainly focuses on the known discrete points varying from 1\% to 20\%. From Fig. \ref{fig:whiteratio}, we can see that even if only 1\% random discrete points remain, the overall of the image can be restored. When 4\%-6\% of the known discrete points remain, the network can make a good restoration. When the proportion of known discrete points exceeds 10\%, even tiny local details of the image can be recovered.

\subsection{Recovery from color-noise}
\begin{figure}
\includegraphics[width=3.2in]{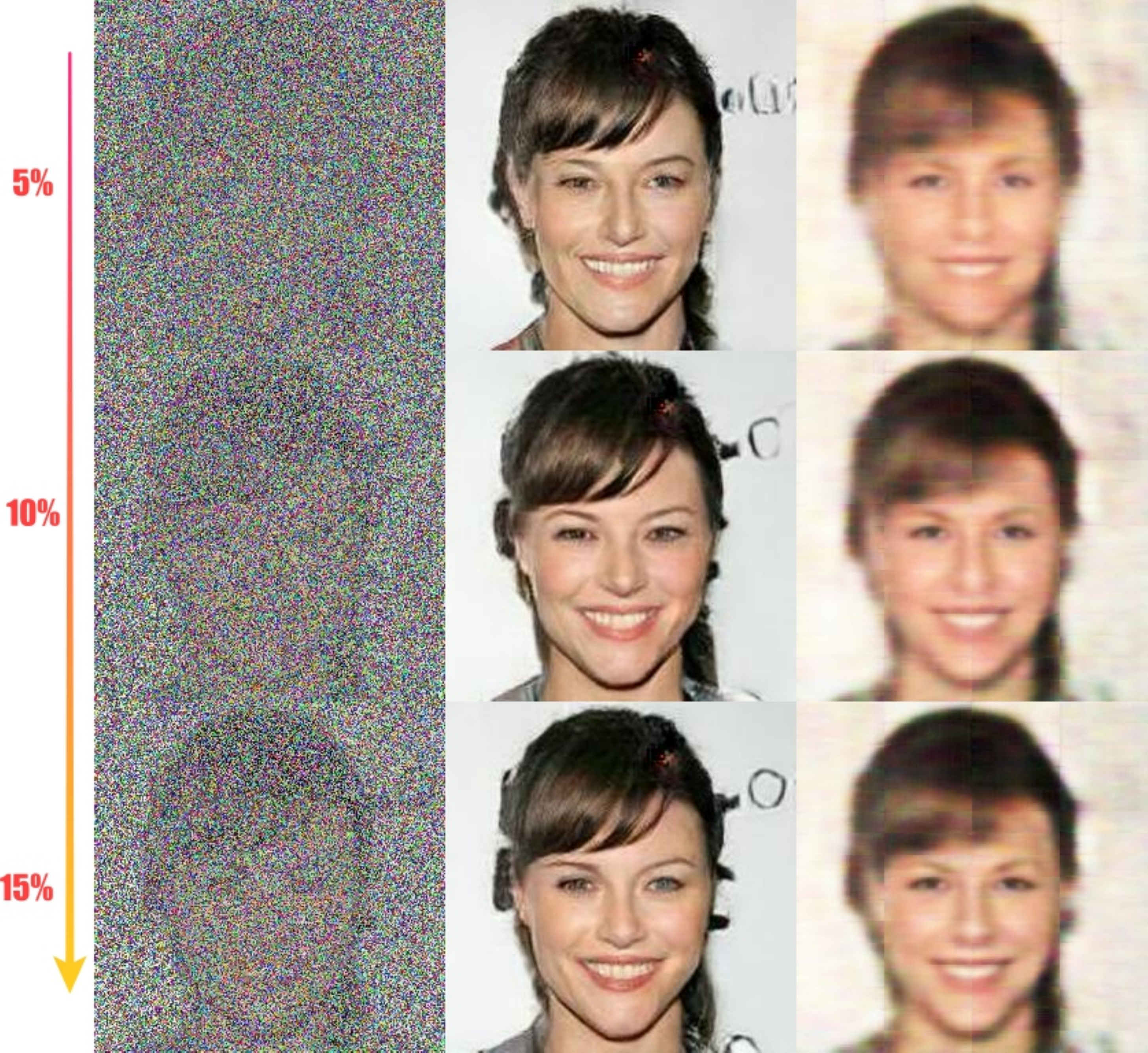}
\caption{Reconstruction of human face from random color noise image from CelebA dataset (valid information varying from 5\%$-$15\%, i.e. color noise varying from 95\%$-$85\%). The first column is the random color noise as the source image, the second column is generated by our method, and the third column is generated by \cite{gao2017demand}. Our method can synthesize better results.}
\label{fig:noiserecoveryface}
\end{figure}

In Fig. \ref{fig:noiserecoveryface}, we demonstrate the recovery from color-noise image of our method in comparison with \cite{gao2017demand}. For the original image containing a large amount of random color noise, our network exhibits good denoising performance. Even when the random color noise accounts for 95\%, we can restore the main content of the image. When the noise drops to 90\%, some details gradually show up. When the noise continues to drop to 85\%, the details of the character's face can be restored very well.

\begin{figure}
\includegraphics[width=3.2in]{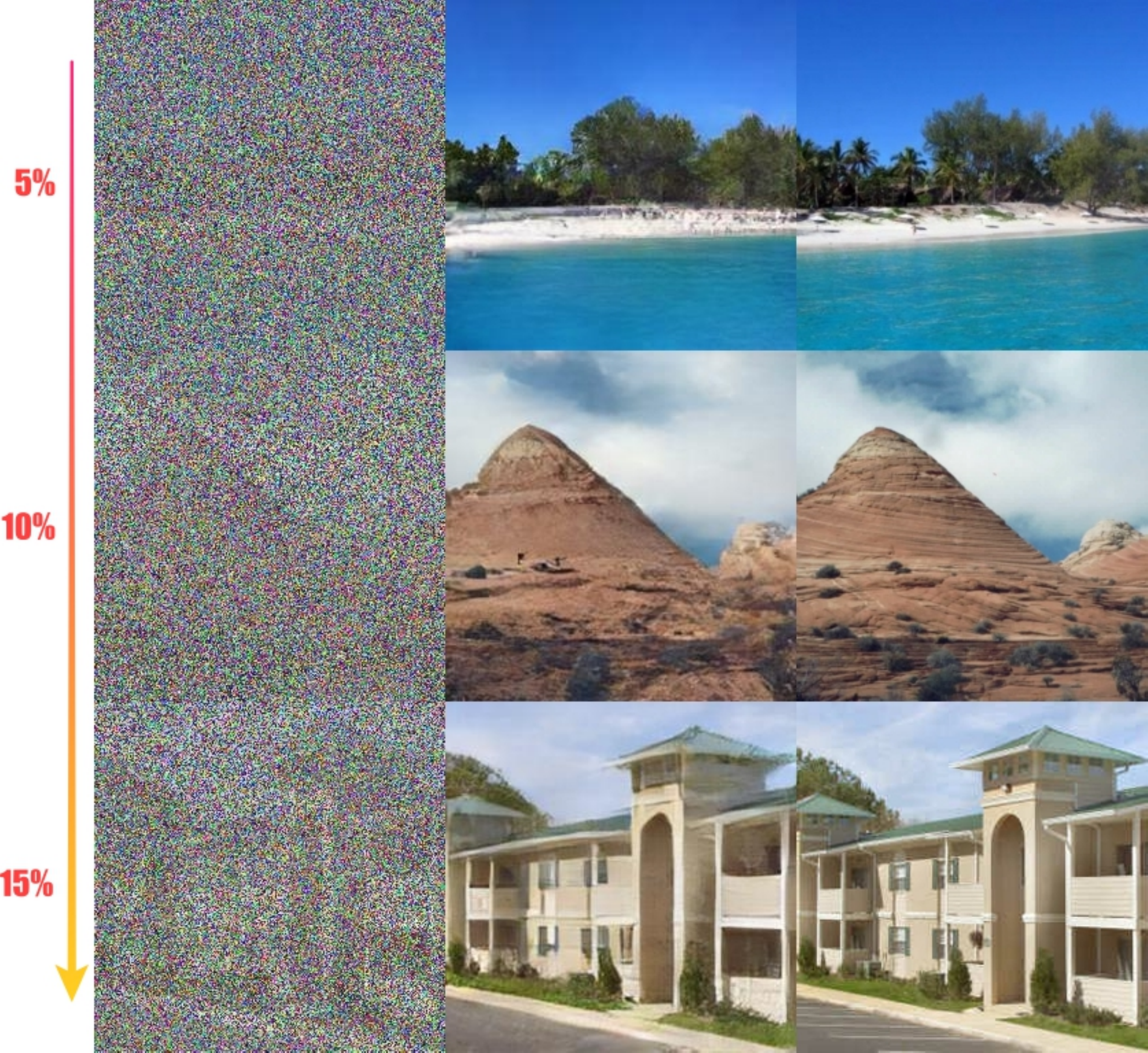}
\caption{Reconstruction of natural scene from random color noise image from SUN397 dataset (valid information varying from 5\%$-$15\%, i.e. color noise varying from 95\%$-$85\%). The first column is the random color noise as the source image, the second column is generated by our method, and the third column is the ground truth. Our method can synthesize good results.}
\label{fig:noiserecoveryscene}
\end{figure}
In Fig. \ref{fig:noiserecoveryscene}, we also demonstrate the recovery from color-noise image of natural scene by our method. We can see our synthesized results are close to the ground truth.

\subsection{Recovery from missing block}
\begin{figure}
\includegraphics[width=3.3in]{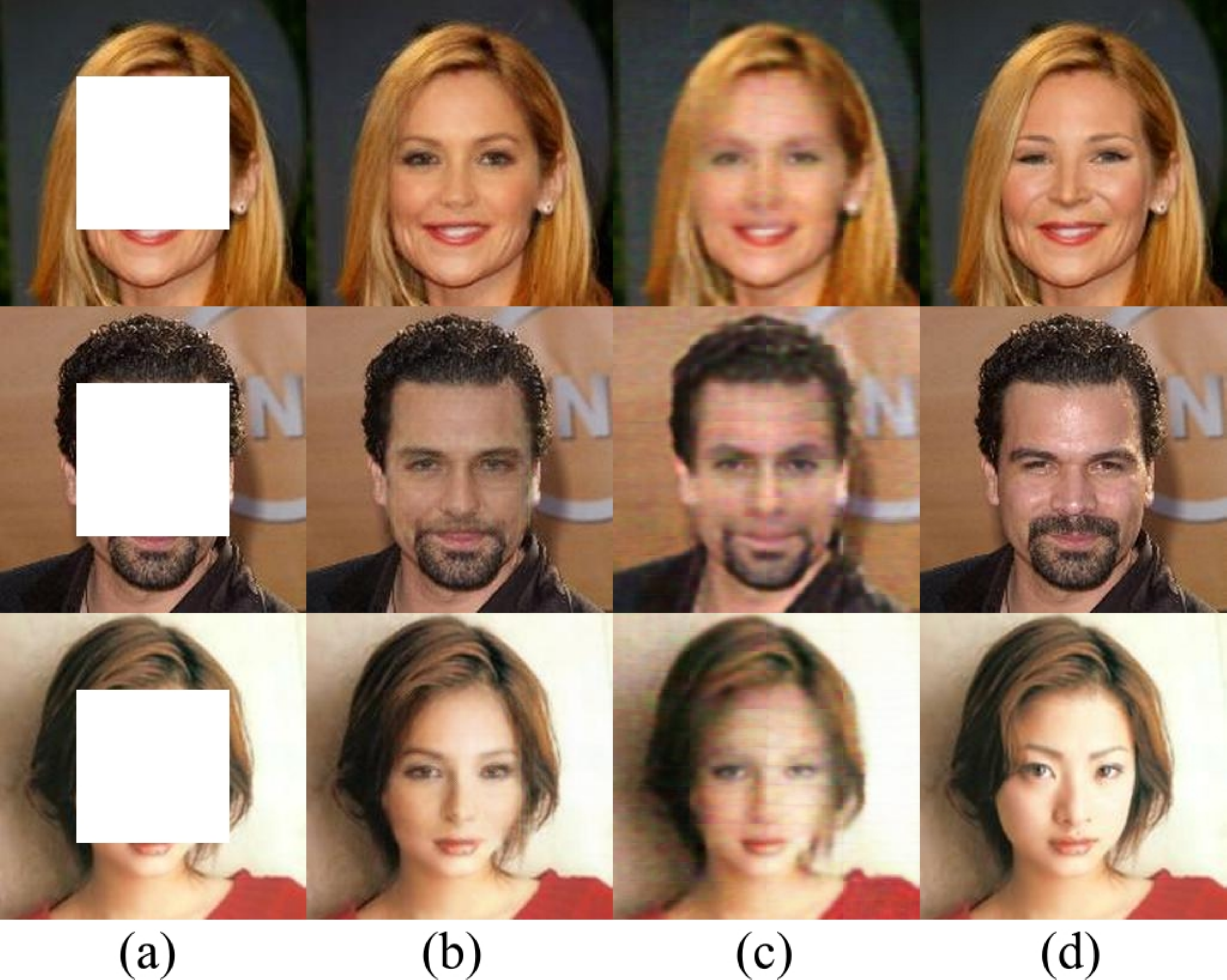}
\caption{Reconstruction from missing block. Column (a) shows the missing block as the source image, column (b) is generated by our method, column (c) is generated by \cite{gao2017demand}, and column (d) is the ground truth. Our method can synthesize better results.}
\label{fig:blockrecovery}
\end{figure}
In fig. \ref{fig:blockrecovery}, we demonstrate the reconstruction from the missing block.
For the missing block in a image (exhibited by white block or constant color block), our method can achieve good inpainting result. The overall of image generated looks harmonious, the character's expression is vivid and can fit closely to the ground truth. The skin color and the pose of the face also match the ground truth well. The illumination of the face is also consistent with the surrounding environment.

\subsection{Recovery from cluttered block}
\begin{figure}
\includegraphics[width=3.4in]{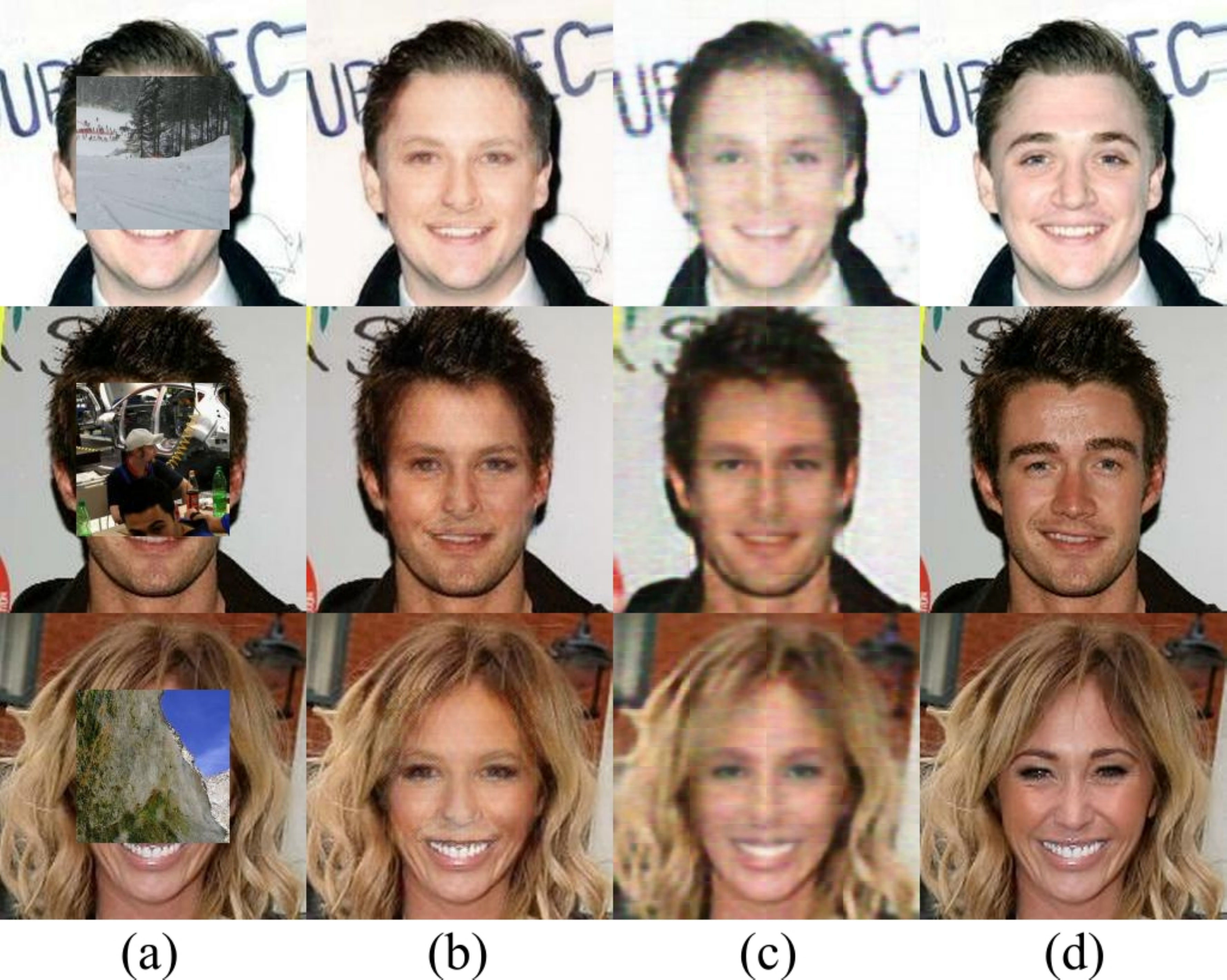}
\caption{Reconstruction from cluttered color block. The first column (a) is the cluttered color block as the source image, column (b) is generated by our method, column (c) is generated by \cite{gao2017demand}, and column (d) is the ground truth. We highlight the better results by our method.}
\label{fig:colorblockrecovery}
\end{figure}

If the missing block is replaced with a cluttered color block, the network not only needs to recover the original information of a image, but also to determine which part of the image should be restored. This increases the difficulty on the network. We highlight our recovery of the image with cluttered block in Fig. \ref{fig:colorblockrecovery}.

\subsection{Extension}
\label{subsection_feature}
\begin{figure}[]
\includegraphics[width=3.2in]{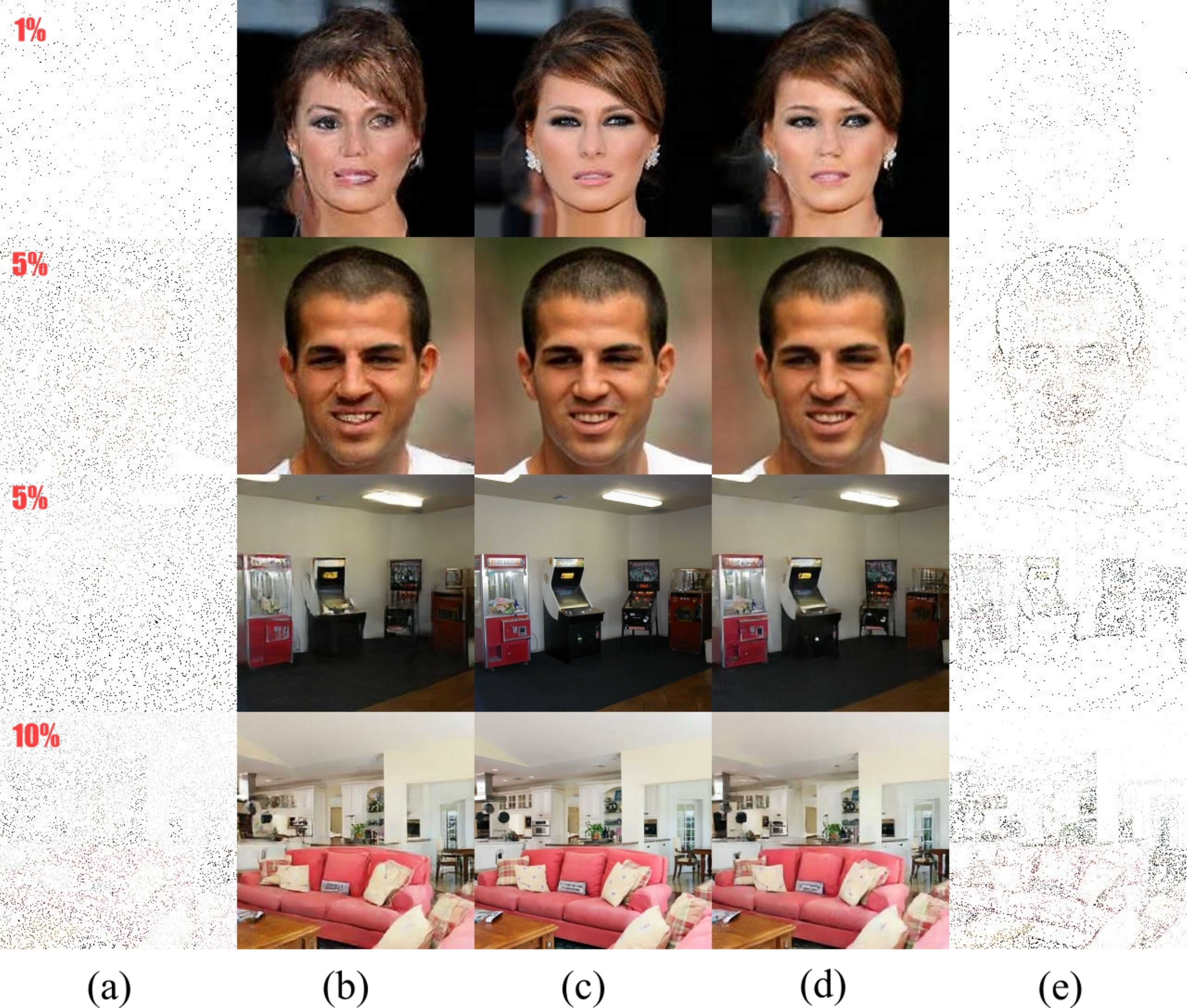}
\caption{Various benchmarks for image reconstruction from very sparse sampling. (a) row shows the random sampling of source image and its reconstruction by our method in (b), (c) row shows the real images, (d) row shows the reconstruction from feature-based sampling of original image in (e). }
\label{fig:sobelcomparison}
\end{figure}

Our network can also be extended to handle image compression and transmission. To realize high-ratio image compression, the samples of a image should be as sparse as possible. Our method can rebuild the image and obtain tolerable result from only 1\% sample points of the original image (see Fig. \ref{fig:sobelcomparison}). To improve the quality of reconstruction, we also propose a sampling strategy according to the cue of edges in the image, which can be considered as important features that should be carefully preserved. We show reconstruction results from very sparse source and compare to the real images in Fig. \ref{fig:sobelcomparison}. From this figure, we can see that scattering more sampling points around the edge results in better reconstruction, and is able to recover fine details of the original image.

\section{Quantitative Comparisons and Analysis}

\begin{table}[]
\centering
\caption{Comparison of reconstruction quality. Our method always has higher PSNR and SSIM value than Gao et.al \cite{gao2017demand}. }
\label{my-label}
\begin{tabular}{c|cc|cc}
\hline
\multirow{2}{*}{} & \multicolumn{2}{c|}{Gao et.al \cite{gao2017demand}} & \multicolumn{2}{c}{Ours} \\ \cline{2-5}
                  & PSNR                 & SSIM\cite{wang2004image}             & PSNR         & SSIM\cite{wang2004image}      \\ \hline
White Block 128x128        & 21.88dB              & 0.68             & 24.99dB      & 0.85      \\
Color Block 128x128        & 21.83dB              & 0.67             & 23.16dB      & 0.82      \\ \hline
\end{tabular}
\label{PSNR}
\end{table}

\subsection{Quantitative comparison}
In Table \ref{PSNR}, we make quantitative comparison with other methods to evaluate the quality of reconstruction with missing white block or cluttered color block(128x128). Image reconstruction is tested on CelebA dataset as the source, and the data of cluttered block come from SUN397 dataset. Our method obtains higher value on both measures of PSNR and SSIM. It implies that our quality of reconstruction is better than the opponent.

\begin{figure}[]
\includegraphics[width=3.2in]{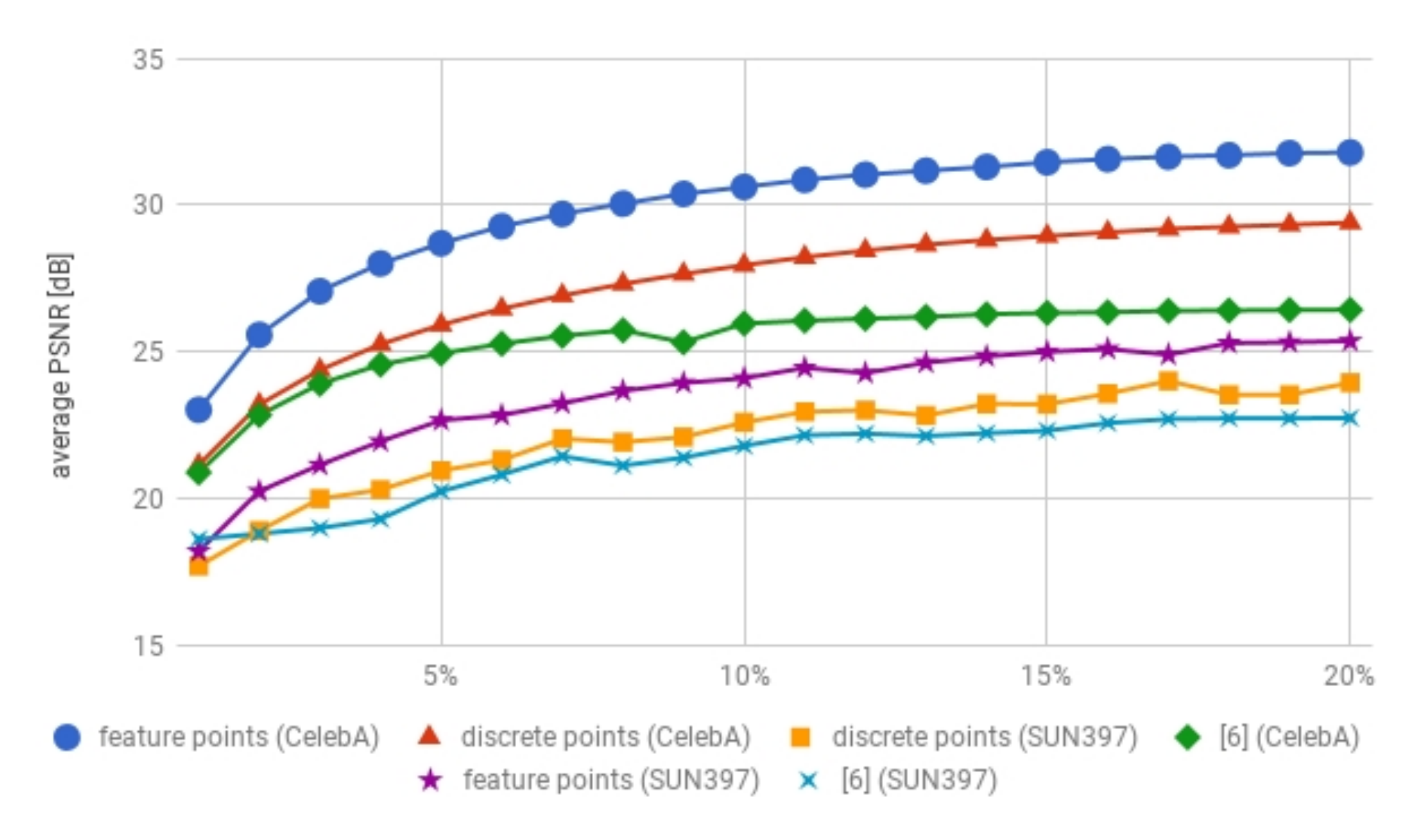}
\caption{Comparisons of reconstruction quality using two different types of datasets: CelebA and SUN397. The reconstruction is conducted using discrete samples and the quality is evaluated using PSNR.}
\label{fig:compareplot}
\end{figure}
The performance of  the reconstruction from a discrete point set is shown in Fig. \ref{fig:compareplot}, where the "feature points" represents the extra processing on point distribution as discussed in subsection \ref{subsection_feature}. We can see that the overall quality of reconstruction using
CelebA dataset is better than using SUN397. This is because the CelebA is specific for human face while SUN397 is more general on the contrary. The reconstruction using the feature point generally has the best quality when compared to those images restored from the random discrete points or the results from \cite{gao2017demand}.

\begin{figure}[]
\includegraphics[width=3.2in]{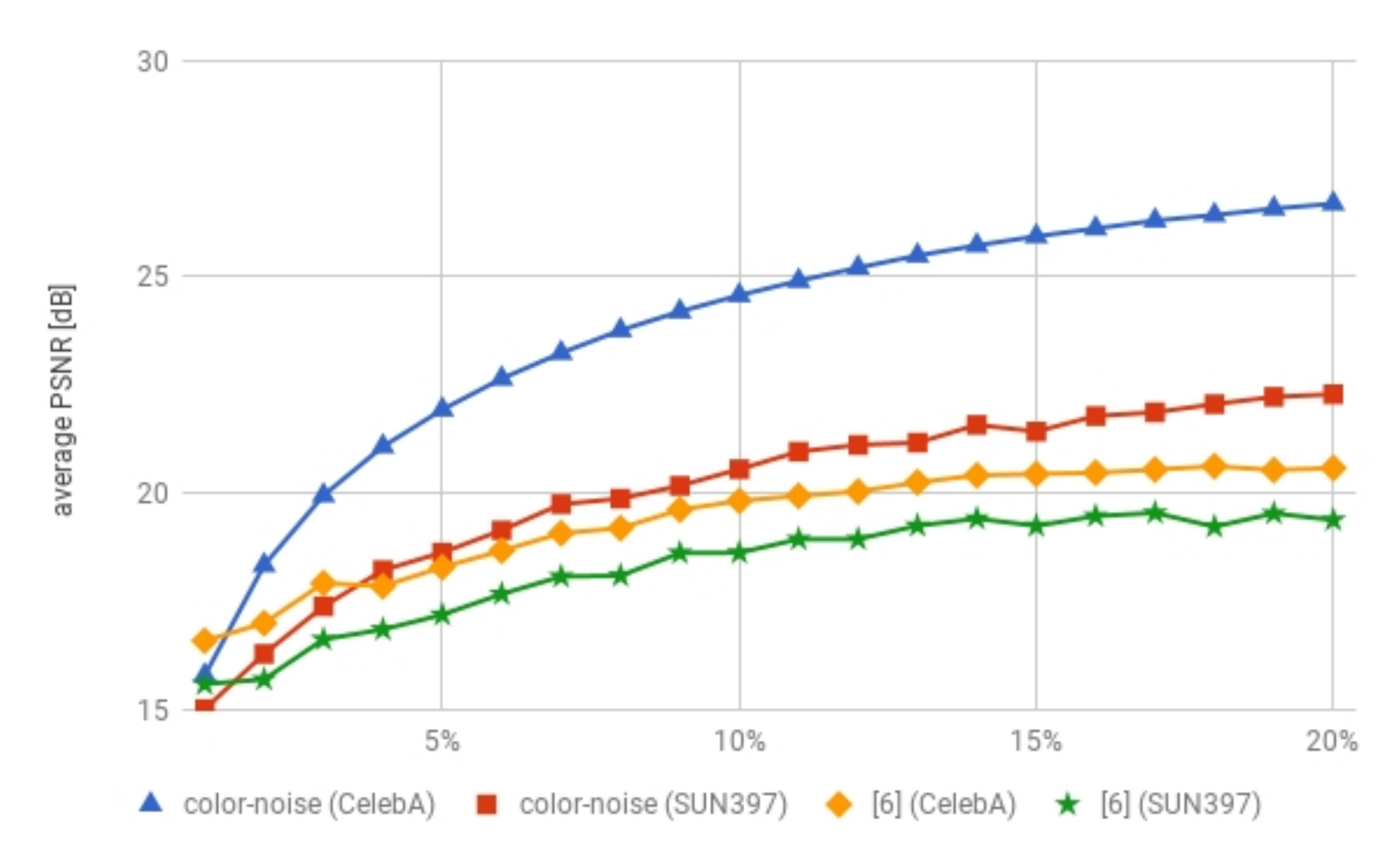}
\caption{For the color-noise task, Our network has a distinct advantage over Gao et.al \cite{gao2017demand}'s network in both CelebA and SUN397 datasets.}
\label{fig:compareplotcolornoise}
\end{figure}

We also compare different approaches for the reconstruction quality when dealing with color noise images in Fig. \ref{fig:compareplotcolornoise}. Our method can also obtain good results.

\begin{figure}[]
\includegraphics[width=3.2in]{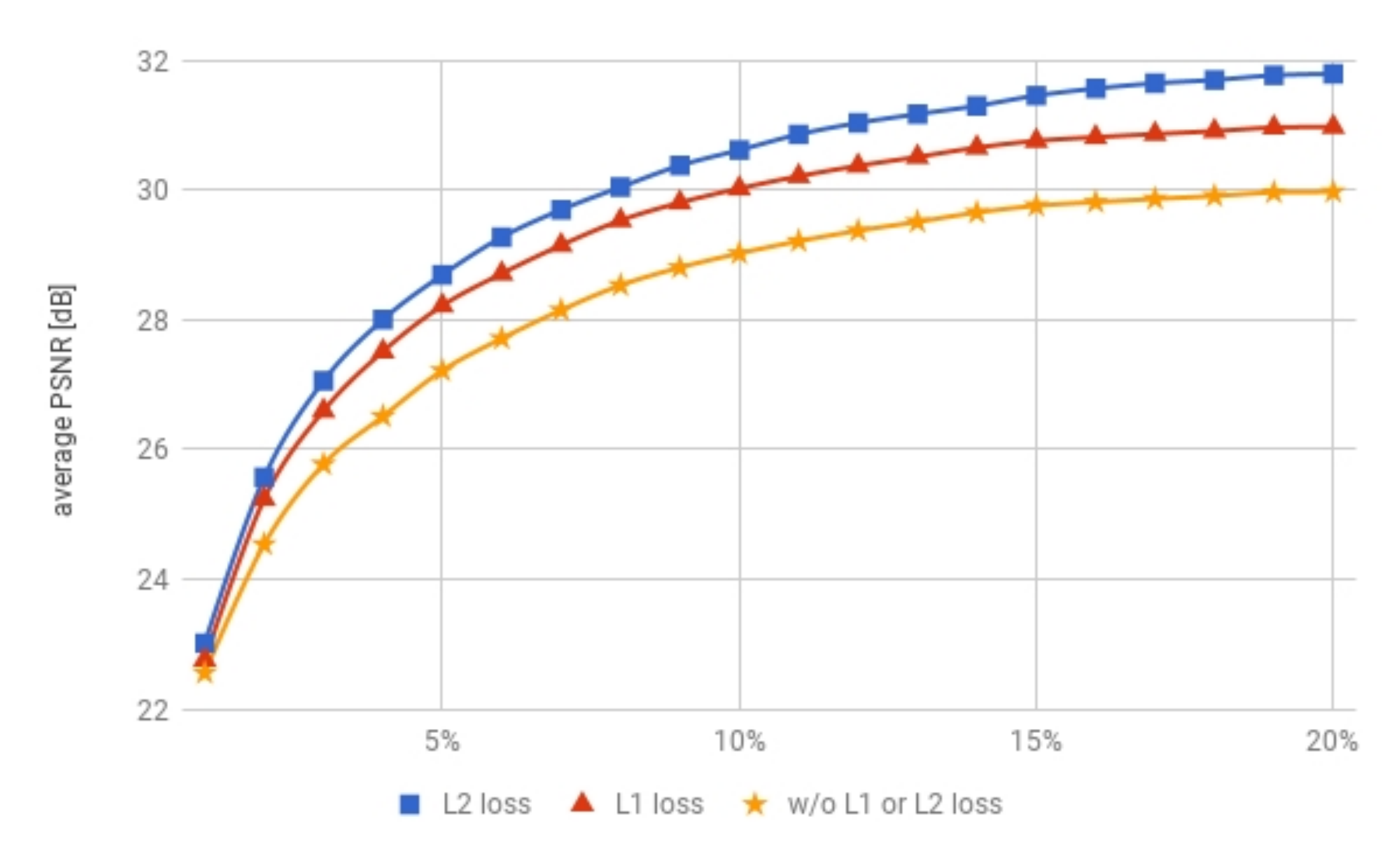}
\caption{By comparing the PSNR values of the images restored at different percentages for each loss function, we find that L2 loss is the best.}
\label{fig:compareplotL1L2}
\end{figure}

For corresponding point loss we compare the strategies with L2 loss, L1 loss, and neither of them. The result in Fig. \ref{fig:compareplotL1L2} shows that L2 loss is the best one among them.

\subsection{Analysis}
The generalization performance of the GANs model addressed in this paper is still restricted by the types of scene, the noise model, and the size of images. That is, different scenes, noise levels, or image sizes may more or less degrade the performance.
We are attempting to use the same random training points as the ones in \cite{gao2017demand} to prevent the trained network from producing common artifacts of deep networks like the checkerboard artifacts. In addition, to achieve the best reduction of small-scale data, discrete sampling points should be scattered in parts of people's attention, such as faces in the image. Saliency maps can also be used for region-of-interests detection, so that the image can be processed further for compression.

\subsection{Discussion}

\textbf{Stability of generated images:} In the GANs, stability is a problem that cannot be ignored. The use of multi-dimensional loss function design to improve the stability of the generated image is a promising solution. The four losses addressed in this paper, namely, the loss calculated by multi-scale discriminators, the adversarial matching loss, the VGG perceptual loss, and the corresponding point loss, describe different perspectives of an image and complement each other.

\textbf{Network structure design:} The input image is of critical importance for a neural network to obtain information. Designing a network that is capable of catch main characteristics of the input image is a good choice. It is out of this consideration that we add corresponding point loss for image recovery.

\textbf{Processed image:} In order to improve the network's ability to recover as many data as possible, it is a good idea to provide more relative information to the network. The scheme of using Sobel operation to improve the performance of image recovery may be a practical way of implementation.

\section{Conclusion}
We present that the conditional GANs based multi-dimensional loss functions network has the ability to solve several branches of image reconstruction, including image restoration, denoising, and inpanting. We show that this framework can restore images with very few known sampling points and obtain good results. We also explore such images with high color noise or cluttered color blocks. Through a certain training network, we can restore the original information in high-noise images. Furthermore, for common image inpanting tasks executed by GANs, we increases the difficulty of training by converting white blocks into cluttered color blocks to explore the network's resilience. Finally, we design a new loss function to help generating more realistic image from badly corrupted images.

\bibliographystyle{ACM-Reference-Format}
\bibliography{Image Reconstruction GANs.bbl}


\begin{thebibliography}{34}


\ifx \showCODEN    \undefined \def \showCODEN     #1{\unskip}     \fi
\ifx \showDOI      \undefined \def \showDOI       #1{#1}\fi
\ifx \showISBNx    \undefined \def \showISBNx     #1{\unskip}     \fi
\ifx \showISBNxiii \undefined \def \showISBNxiii  #1{\unskip}     \fi
\ifx \showISSN     \undefined \def \showISSN      #1{\unskip}     \fi
\ifx \showLCCN     \undefined \def \showLCCN      #1{\unskip}     \fi
\ifx \shownote     \undefined \def \shownote      #1{#1}          \fi
\ifx \showarticletitle \undefined \def \showarticletitle #1{#1}   \fi
\ifx \showURL      \undefined \def \showURL       {\relax}        \fi
\providecommand\bibfield[2]{#2}
\providecommand\bibinfo[2]{#2}
\providecommand\natexlab[1]{#1}
\providecommand\showeprint[2][]{arXiv:#2}

\bibitem[\protect\citeauthoryear{Burger, Schuler, and Harmeling}{Burger
  et~al\mbox{.}}{2012}]%
        {burger2012image}
\bibfield{author}{\bibinfo{person}{Harold~C Burger},
  \bibinfo{person}{Christian~J Schuler}, {and} \bibinfo{person}{Stefan
  Harmeling}.} \bibinfo{year}{2012}\natexlab{}.
\newblock \showarticletitle{Image denoising: Can plain neural networks compete
  with BM3D?}. In \bibinfo{booktitle}{\emph{Computer Vision and Pattern
  Recognition (CVPR), 2012 IEEE Conference on}}. IEEE,
  \bibinfo{pages}{2392--2399}.
\newblock


\bibitem[\protect\citeauthoryear{Cho, Tai, and Kweon}{Cho
  et~al\mbox{.}}{2016}]%
        {cho2016natural}
\bibfield{author}{\bibinfo{person}{Donghyeon Cho}, \bibinfo{person}{Yu-Wing
  Tai}, {and} \bibinfo{person}{Inso Kweon}.} \bibinfo{year}{2016}\natexlab{}.
\newblock \showarticletitle{Natural image matting using deep convolutional
  neural networks}. In \bibinfo{booktitle}{\emph{European Conference on
  Computer Vision}}. Springer, \bibinfo{pages}{626--643}.
\newblock


\bibitem[\protect\citeauthoryear{Dong, Loy, He, and Tang}{Dong
  et~al\mbox{.}}{2016}]%
        {dong2016image}
\bibfield{author}{\bibinfo{person}{Chao Dong}, \bibinfo{person}{Chen~Change
  Loy}, \bibinfo{person}{Kaiming He}, {and} \bibinfo{person}{Xiaoou Tang}.}
  \bibinfo{year}{2016}\natexlab{}.
\newblock \showarticletitle{Image super-resolution using deep convolutional
  networks}.
\newblock \bibinfo{journal}{\emph{IEEE transactions on pattern analysis and
  machine intelligence}} \bibinfo{volume}{38}, \bibinfo{number}{2}
  (\bibinfo{year}{2016}), \bibinfo{pages}{295--307}.
\newblock


\bibitem[\protect\citeauthoryear{Dosovitskiy and Brox}{Dosovitskiy and
  Brox}{2016}]%
        {dosovitskiy2016generating}
\bibfield{author}{\bibinfo{person}{Alexey Dosovitskiy} {and}
  \bibinfo{person}{Thomas Brox}.} \bibinfo{year}{2016}\natexlab{}.
\newblock \showarticletitle{Generating images with perceptual similarity
  metrics based on deep networks}. In \bibinfo{booktitle}{\emph{Advances in
  Neural Information Processing Systems}}. \bibinfo{pages}{658--666}.
\newblock


\bibitem[\protect\citeauthoryear{Durugkar, Gemp, and Mahadevan}{Durugkar
  et~al\mbox{.}}{2016}]%
        {durugkar2016generative}
\bibfield{author}{\bibinfo{person}{Ishan Durugkar}, \bibinfo{person}{Ian Gemp},
  {and} \bibinfo{person}{Sridhar Mahadevan}.} \bibinfo{year}{2016}\natexlab{}.
\newblock \showarticletitle{Generative multi-adversarial networks}.
\newblock \bibinfo{journal}{\emph{arXiv preprint arXiv:1611.01673}}
  (\bibinfo{year}{2016}).
\newblock


\bibitem[\protect\citeauthoryear{Gao and Grauman}{Gao and Grauman}{2017}]%
        {gao2017demand}
\bibfield{author}{\bibinfo{person}{Ruohan Gao} {and} \bibinfo{person}{Kristen
  Grauman}.} \bibinfo{year}{2017}\natexlab{}.
\newblock \showarticletitle{On-demand learning for deep image restoration}. In
  \bibinfo{booktitle}{\emph{Proc. IEEE Conf. Comput. Vision and Pattern
  Recognition}}. \bibinfo{pages}{1086--1095}.
\newblock


\bibitem[\protect\citeauthoryear{Gatys, Ecker, and Bethge}{Gatys
  et~al\mbox{.}}{2016}]%
        {gatys2016image}
\bibfield{author}{\bibinfo{person}{Leon~A Gatys}, \bibinfo{person}{Alexander~S
  Ecker}, {and} \bibinfo{person}{Matthias Bethge}.}
  \bibinfo{year}{2016}\natexlab{}.
\newblock \showarticletitle{Image style transfer using convolutional neural
  networks}. In \bibinfo{booktitle}{\emph{Computer Vision and Pattern
  Recognition (CVPR), 2016 IEEE Conference on}}. IEEE,
  \bibinfo{pages}{2414--2423}.
\newblock


\bibitem[\protect\citeauthoryear{Goodfellow, Pouget-Abadie, Mirza, Xu,
  Warde-Farley, Ozair, Courville, and Bengio}{Goodfellow et~al\mbox{.}}{2014}]%
        {goodfellow2014generative}
\bibfield{author}{\bibinfo{person}{Ian Goodfellow}, \bibinfo{person}{Jean
  Pouget-Abadie}, \bibinfo{person}{Mehdi Mirza}, \bibinfo{person}{Bing Xu},
  \bibinfo{person}{David Warde-Farley}, \bibinfo{person}{Sherjil Ozair},
  \bibinfo{person}{Aaron Courville}, {and} \bibinfo{person}{Yoshua Bengio}.}
  \bibinfo{year}{2014}\natexlab{}.
\newblock \showarticletitle{Generative adversarial nets}. In
  \bibinfo{booktitle}{\emph{Advances in neural information processing
  systems}}. \bibinfo{pages}{2672--2680}.
\newblock


\bibitem[\protect\citeauthoryear{He, Zhang, Ren, and Sun}{He
  et~al\mbox{.}}{2016}]%
        {he2016deep}
\bibfield{author}{\bibinfo{person}{Kaiming He}, \bibinfo{person}{Xiangyu
  Zhang}, \bibinfo{person}{Shaoqing Ren}, {and} \bibinfo{person}{Jian Sun}.}
  \bibinfo{year}{2016}\natexlab{}.
\newblock \showarticletitle{Deep residual learning for image recognition}. In
  \bibinfo{booktitle}{\emph{Proceedings of the IEEE conference on computer
  vision and pattern recognition}}. \bibinfo{pages}{770--778}.
\newblock


\bibitem[\protect\citeauthoryear{Iizuka, Simo-Serra, and Ishikawa}{Iizuka
  et~al\mbox{.}}{2017}]%
        {iizuka2017globally}
\bibfield{author}{\bibinfo{person}{Satoshi Iizuka}, \bibinfo{person}{Edgar
  Simo-Serra}, {and} \bibinfo{person}{Hiroshi Ishikawa}.}
  \bibinfo{year}{2017}\natexlab{}.
\newblock \showarticletitle{Globally and locally consistent image completion}.
\newblock \bibinfo{journal}{\emph{ACM Transactions on Graphics (TOG)}}
  \bibinfo{volume}{36}, \bibinfo{number}{4} (\bibinfo{year}{2017}),
  \bibinfo{pages}{107}.
\newblock


\bibitem[\protect\citeauthoryear{Ioffe and Szegedy}{Ioffe and Szegedy}{2015}]%
        {ioffe2015batch}
\bibfield{author}{\bibinfo{person}{Sergey Ioffe} {and}
  \bibinfo{person}{Christian Szegedy}.} \bibinfo{year}{2015}\natexlab{}.
\newblock \showarticletitle{Batch normalization: Accelerating deep network
  training by reducing internal covariate shift}.
\newblock \bibinfo{journal}{\emph{arXiv preprint arXiv:1502.03167}}
  (\bibinfo{year}{2015}).
\newblock


\bibitem[\protect\citeauthoryear{Isola, Zhu, Zhou, and Efros}{Isola
  et~al\mbox{.}}{2017}]%
        {isola2017image}
\bibfield{author}{\bibinfo{person}{Phillip Isola}, \bibinfo{person}{Jun-Yan
  Zhu}, \bibinfo{person}{Tinghui Zhou}, {and} \bibinfo{person}{Alexei~A
  Efros}.} \bibinfo{year}{2017}\natexlab{}.
\newblock \showarticletitle{Image-to-image translation with conditional
  adversarial networks}.
\newblock \bibinfo{journal}{\emph{arXiv preprint}} (\bibinfo{year}{2017}).
\newblock


\bibitem[\protect\citeauthoryear{Johnson, Alahi, and Fei-Fei}{Johnson
  et~al\mbox{.}}{2016}]%
        {johnson2016perceptual}
\bibfield{author}{\bibinfo{person}{Justin Johnson}, \bibinfo{person}{Alexandre
  Alahi}, {and} \bibinfo{person}{Li Fei-Fei}.} \bibinfo{year}{2016}\natexlab{}.
\newblock \showarticletitle{Perceptual losses for real-time style transfer and
  super-resolution}. In \bibinfo{booktitle}{\emph{European Conference on
  Computer Vision}}. Springer, \bibinfo{pages}{694--711}.
\newblock


\bibitem[\protect\citeauthoryear{Kingma and Ba}{Kingma and Ba}{2014}]%
        {kingma2014adam}
\bibfield{author}{\bibinfo{person}{Diederik~P Kingma} {and}
  \bibinfo{person}{Jimmy Ba}.} \bibinfo{year}{2014}\natexlab{}.
\newblock \showarticletitle{Adam: A method for stochastic optimization}.
\newblock \bibinfo{journal}{\emph{arXiv preprint arXiv:1412.6980}}
  (\bibinfo{year}{2014}).
\newblock


\bibitem[\protect\citeauthoryear{Krizhevsky, Sutskever, and Hinton}{Krizhevsky
  et~al\mbox{.}}{2012}]%
        {krizhevsky2012imagenet}
\bibfield{author}{\bibinfo{person}{Alex Krizhevsky}, \bibinfo{person}{Ilya
  Sutskever}, {and} \bibinfo{person}{Geoffrey~E Hinton}.}
  \bibinfo{year}{2012}\natexlab{}.
\newblock \showarticletitle{Imagenet classification with deep convolutional
  neural networks}. In \bibinfo{booktitle}{\emph{Advances in neural information
  processing systems}}. \bibinfo{pages}{1097--1105}.
\newblock


\bibitem[\protect\citeauthoryear{Larsson, Maire, and Shakhnarovich}{Larsson
  et~al\mbox{.}}{2016}]%
        {larsson2016learning}
\bibfield{author}{\bibinfo{person}{Gustav Larsson}, \bibinfo{person}{Michael
  Maire}, {and} \bibinfo{person}{Gregory Shakhnarovich}.}
  \bibinfo{year}{2016}\natexlab{}.
\newblock \showarticletitle{Learning representations for automatic
  colorization}. In \bibinfo{booktitle}{\emph{European Conference on Computer
  Vision}}. Springer, \bibinfo{pages}{577--593}.
\newblock


\bibitem[\protect\citeauthoryear{Ledig, Theis, Husz{\'a}r, Caballero,
  Cunningham, Acosta, Aitken, Tejani, Totz, Wang, et~al\mbox{.}}{Ledig
  et~al\mbox{.}}{2016}]%
        {ledig2016photo}
\bibfield{author}{\bibinfo{person}{Christian Ledig}, \bibinfo{person}{Lucas
  Theis}, \bibinfo{person}{Ferenc Husz{\'a}r}, \bibinfo{person}{Jose
  Caballero}, \bibinfo{person}{Andrew Cunningham}, \bibinfo{person}{Alejandro
  Acosta}, \bibinfo{person}{Andrew Aitken}, \bibinfo{person}{Alykhan Tejani},
  \bibinfo{person}{Johannes Totz}, \bibinfo{person}{Zehan Wang},
  {et~al\mbox{.}}} \bibinfo{year}{2016}\natexlab{}.
\newblock \showarticletitle{Photo-realistic single image super-resolution using
  a generative adversarial network}.
\newblock \bibinfo{journal}{\emph{arXiv preprint}} (\bibinfo{year}{2016}).
\newblock


\bibitem[\protect\citeauthoryear{Levin, Lischinski, and Weiss}{Levin
  et~al\mbox{.}}{2004}]%
        {levin2004colorization}
\bibfield{author}{\bibinfo{person}{Anat Levin}, \bibinfo{person}{Dani
  Lischinski}, {and} \bibinfo{person}{Yair Weiss}.}
  \bibinfo{year}{2004}\natexlab{}.
\newblock \showarticletitle{Colorization using optimization}. In
  \bibinfo{booktitle}{\emph{ACM Transactions on Graphics (ToG)}},
  Vol.~\bibinfo{volume}{23}. ACM, \bibinfo{pages}{689--694}.
\newblock


\bibitem[\protect\citeauthoryear{Levin, Lischinski, and Weiss}{Levin
  et~al\mbox{.}}{2008}]%
        {levin2008closed}
\bibfield{author}{\bibinfo{person}{Anat Levin}, \bibinfo{person}{Dani
  Lischinski}, {and} \bibinfo{person}{Yair Weiss}.}
  \bibinfo{year}{2008}\natexlab{}.
\newblock \showarticletitle{A closed-form solution to natural image matting}.
\newblock \bibinfo{journal}{\emph{IEEE Transactions on Pattern Analysis and
  Machine Intelligence}} \bibinfo{volume}{30}, \bibinfo{number}{2}
  (\bibinfo{year}{2008}), \bibinfo{pages}{228--242}.
\newblock


\bibitem[\protect\citeauthoryear{Liu, Pan, and Yang}{Liu et~al\mbox{.}}{2016}]%
        {liu2016learning}
\bibfield{author}{\bibinfo{person}{Sifei Liu}, \bibinfo{person}{Jinshan Pan},
  {and} \bibinfo{person}{Ming-Hsuan Yang}.} \bibinfo{year}{2016}\natexlab{}.
\newblock \showarticletitle{Learning recursive filters for low-level vision via
  a hybrid neural network}. In \bibinfo{booktitle}{\emph{European Conference on
  Computer Vision}}. Springer, \bibinfo{pages}{560--576}.
\newblock


\bibitem[\protect\citeauthoryear{Liu, Luo, Wang, and Tang}{Liu
  et~al\mbox{.}}{2015}]%
        {liu2015deep}
\bibfield{author}{\bibinfo{person}{Ziwei Liu}, \bibinfo{person}{Ping Luo},
  \bibinfo{person}{Xiaogang Wang}, {and} \bibinfo{person}{Xiaoou Tang}.}
  \bibinfo{year}{2015}\natexlab{}.
\newblock \showarticletitle{Deep learning face attributes in the wild}. In
  \bibinfo{booktitle}{\emph{Proceedings of the IEEE International Conference on
  Computer Vision}}. \bibinfo{pages}{3730--3738}.
\newblock


\bibitem[\protect\citeauthoryear{Long, Shelhamer, and Darrell}{Long
  et~al\mbox{.}}{2015}]%
        {long2015fully}
\bibfield{author}{\bibinfo{person}{Jonathan Long}, \bibinfo{person}{Evan
  Shelhamer}, {and} \bibinfo{person}{Trevor Darrell}.}
  \bibinfo{year}{2015}\natexlab{}.
\newblock \showarticletitle{Fully convolutional networks for semantic
  segmentation}. In \bibinfo{booktitle}{\emph{Proceedings of the IEEE
  conference on computer vision and pattern recognition}}.
  \bibinfo{pages}{3431--3440}.
\newblock


\bibitem[\protect\citeauthoryear{Mirza and Osindero}{Mirza and
  Osindero}{2014}]%
        {mirza2014conditional}
\bibfield{author}{\bibinfo{person}{Mehdi Mirza} {and} \bibinfo{person}{Simon
  Osindero}.} \bibinfo{year}{2014}\natexlab{}.
\newblock \showarticletitle{Conditional generative adversarial nets}.
\newblock \bibinfo{journal}{\emph{arXiv preprint arXiv:1411.1784}}
  (\bibinfo{year}{2014}).
\newblock


\bibitem[\protect\citeauthoryear{Noh, Hong, and Han}{Noh et~al\mbox{.}}{2015}]%
        {noh2015learning}
\bibfield{author}{\bibinfo{person}{Hyeonwoo Noh}, \bibinfo{person}{Seunghoon
  Hong}, {and} \bibinfo{person}{Bohyung Han}.} \bibinfo{year}{2015}\natexlab{}.
\newblock \showarticletitle{Learning deconvolution network for semantic
  segmentation}. In \bibinfo{booktitle}{\emph{Proceedings of the IEEE
  International Conference on Computer Vision}}. \bibinfo{pages}{1520--1528}.
\newblock


\bibitem[\protect\citeauthoryear{Pathak, Krahenbuhl, Donahue, Darrell, and
  Efros}{Pathak et~al\mbox{.}}{2016}]%
        {pathak2016context}
\bibfield{author}{\bibinfo{person}{Deepak Pathak}, \bibinfo{person}{Philipp
  Krahenbuhl}, \bibinfo{person}{Jeff Donahue}, \bibinfo{person}{Trevor
  Darrell}, {and} \bibinfo{person}{Alexei~A Efros}.}
  \bibinfo{year}{2016}\natexlab{}.
\newblock \showarticletitle{Context encoders: Feature learning by inpainting}.
  In \bibinfo{booktitle}{\emph{Proceedings of the IEEE Conference on Computer
  Vision and Pattern Recognition}}. \bibinfo{pages}{2536--2544}.
\newblock


\bibitem[\protect\citeauthoryear{Simonyan and Zisserman}{Simonyan and
  Zisserman}{2014}]%
        {simonyan2014very}
\bibfield{author}{\bibinfo{person}{Karen Simonyan} {and}
  \bibinfo{person}{Andrew Zisserman}.} \bibinfo{year}{2014}\natexlab{}.
\newblock \showarticletitle{Very deep convolutional networks for large-scale
  image recognition}.
\newblock \bibinfo{journal}{\emph{arXiv preprint arXiv:1409.1556}}
  (\bibinfo{year}{2014}).
\newblock


\bibitem[\protect\citeauthoryear{Sobel}{Sobel}{2014}]%
        {sobel2014history}
\bibfield{author}{\bibinfo{person}{Irwin Sobel}.}
  \bibinfo{year}{2014}\natexlab{}.
\newblock \showarticletitle{History and definition of the sobel operator}.
\newblock \bibinfo{journal}{\emph{Retrieved from the World Wide Web}}
  (\bibinfo{year}{2014}).
\newblock


\bibitem[\protect\citeauthoryear{Wang, Liu, Zhu, Tao, Kautz, and
  Catanzaro}{Wang et~al\mbox{.}}{2017}]%
        {wang2017high}
\bibfield{author}{\bibinfo{person}{Ting-Chun Wang}, \bibinfo{person}{Ming-Yu
  Liu}, \bibinfo{person}{Jun-Yan Zhu}, \bibinfo{person}{Andrew Tao},
  \bibinfo{person}{Jan Kautz}, {and} \bibinfo{person}{Bryan Catanzaro}.}
  \bibinfo{year}{2017}\natexlab{}.
\newblock \showarticletitle{High-Resolution Image Synthesis and Semantic
  Manipulation with Conditional GANs}.
\newblock \bibinfo{journal}{\emph{arXiv preprint arXiv:1711.11585}}
  (\bibinfo{year}{2017}).
\newblock


\bibitem[\protect\citeauthoryear{Wang, Bovik, Sheikh, and Simoncelli}{Wang
  et~al\mbox{.}}{2004}]%
        {wang2004image}
\bibfield{author}{\bibinfo{person}{Zhou Wang}, \bibinfo{person}{Alan~C Bovik},
  \bibinfo{person}{Hamid~R Sheikh}, {and} \bibinfo{person}{Eero~P Simoncelli}.}
  \bibinfo{year}{2004}\natexlab{}.
\newblock \showarticletitle{Image quality assessment: from error visibility to
  structural similarity}.
\newblock \bibinfo{journal}{\emph{IEEE transactions on image processing}}
  \bibinfo{volume}{13}, \bibinfo{number}{4} (\bibinfo{year}{2004}),
  \bibinfo{pages}{600--612}.
\newblock


\bibitem[\protect\citeauthoryear{Xiao, Ehinger, Hays, Torralba, and Oliva}{Xiao
  et~al\mbox{.}}{2016}]%
        {xiao2016sun}
\bibfield{author}{\bibinfo{person}{Jianxiong Xiao}, \bibinfo{person}{Krista~A
  Ehinger}, \bibinfo{person}{James Hays}, \bibinfo{person}{Antonio Torralba},
  {and} \bibinfo{person}{Aude Oliva}.} \bibinfo{year}{2016}\natexlab{}.
\newblock \showarticletitle{Sun database: Exploring a large collection of scene
  categories}.
\newblock \bibinfo{journal}{\emph{International Journal of Computer Vision}}
  \bibinfo{volume}{119}, \bibinfo{number}{1} (\bibinfo{year}{2016}),
  \bibinfo{pages}{3--22}.
\newblock


\bibitem[\protect\citeauthoryear{Xu, Yan, and Jia}{Xu et~al\mbox{.}}{2013}]%
        {xu2013sparse}
\bibfield{author}{\bibinfo{person}{Li Xu}, \bibinfo{person}{Qiong Yan}, {and}
  \bibinfo{person}{Jiaya Jia}.} \bibinfo{year}{2013}\natexlab{}.
\newblock \showarticletitle{A sparse control model for image and video
  editing}.
\newblock \bibinfo{journal}{\emph{ACM Transactions on Graphics (TOG)}}
  \bibinfo{volume}{32}, \bibinfo{number}{6} (\bibinfo{year}{2013}),
  \bibinfo{pages}{197}.
\newblock


\bibitem[\protect\citeauthoryear{Yeh, Chen, Lim, Schwing, Hasegawa-Johnson, and
  Do}{Yeh et~al\mbox{.}}{2017}]%
        {yeh2017semantic}
\bibfield{author}{\bibinfo{person}{Raymond~A Yeh}, \bibinfo{person}{Chen Chen},
  \bibinfo{person}{Teck~Yian Lim}, \bibinfo{person}{Alexander~G Schwing},
  \bibinfo{person}{Mark Hasegawa-Johnson}, {and} \bibinfo{person}{Minh~N Do}.}
  \bibinfo{year}{2017}\natexlab{}.
\newblock \showarticletitle{Semantic image inpainting with deep generative
  models}. In \bibinfo{booktitle}{\emph{Proceedings of the IEEE Conference on
  Computer Vision and Pattern Recognition}}. \bibinfo{pages}{5485--5493}.
\newblock


\bibitem[\protect\citeauthoryear{Zhang, Zuo, Chen, Meng, and Zhang}{Zhang
  et~al\mbox{.}}{2017}]%
        {zhang2017beyond}
\bibfield{author}{\bibinfo{person}{Kai Zhang}, \bibinfo{person}{Wangmeng Zuo},
  \bibinfo{person}{Yunjin Chen}, \bibinfo{person}{Deyu Meng}, {and}
  \bibinfo{person}{Lei Zhang}.} \bibinfo{year}{2017}\natexlab{}.
\newblock \showarticletitle{Beyond a gaussian denoiser: Residual learning of
  deep cnn for image denoising}.
\newblock \bibinfo{journal}{\emph{IEEE Transactions on Image Processing}}
  \bibinfo{volume}{26}, \bibinfo{number}{7} (\bibinfo{year}{2017}),
  \bibinfo{pages}{3142--3155}.
\newblock


\bibitem[\protect\citeauthoryear{Zhu, Kr{\"a}henb{\"u}hl, Shechtman, and
  Efros}{Zhu et~al\mbox{.}}{2016}]%
        {zhu2016generative}
\bibfield{author}{\bibinfo{person}{Jun-Yan Zhu}, \bibinfo{person}{Philipp
  Kr{\"a}henb{\"u}hl}, \bibinfo{person}{Eli Shechtman}, {and}
  \bibinfo{person}{Alexei~A Efros}.} \bibinfo{year}{2016}\natexlab{}.
\newblock \showarticletitle{Generative visual manipulation on the natural image
  manifold}. In \bibinfo{booktitle}{\emph{European Conference on Computer
  Vision}}. Springer, \bibinfo{pages}{597--613}.
\newblock


\end{thebibliography}

\end{document}